# Classifier Fusion Method to Recognize Handwritten Kannada Numerals

Mamatha H.R
Department of ISE
PES Institute of Technology
Bangalore, India
mamatha.bhr@gmail.com

Karthik S
Department of ISE
PES School of Engineering
Bangalore, India
karthiks@pes.edu

Srikanta Murthy K
Department of CSE
PES School of Engineering
Bangalore, India
shree194@gmail.com

*Abstract*— Optical Character Recognition (OCR) is one of the important fields in image processing and pattern recognition domain. Handwritten character recognition has always been a challenging task. Only a little work can be traced towards the recognition of handwritten characters for the south Indian languages. Kannada is one such south Indian language which is also one of the official language of India. Accurate recognition of Kannada characters is a challenging task because of the high degree of similarity between the characters. Hence, good quality features are to be extracted and better classifiers are needed to improve the accuracy of the OCR for Kannada characters. This paper explores the effectiveness of feature extraction method like run length count (RLC) and directional chain code (DCC) for the recognition of handwritten Kannada numerals. In this paper, a classifier fusion method is implemented to improve the recognition rate. For the classifier fusion, we have considered K-nearest neighbour (KNN) and Linear classifier (LC). The novelty of this method is to achieve better accuracy with few features using classifier fusion approach. Proposed method achieves an average recognition rate of 96%.

*Keywords— OCR, handwritten Kannada numeral, directional chain code, run length count, K-Nearest Neighbour, Linear classifier, classifier fusion*

I. INTRODUCTION

The recognition of handwritten numeral is an important area of research for its applications in post office, banks and other organizations. Handwritten character recognition (HCR) has received extensive attention in academic and production fields. The recognition system can be either on-line or off-line. Off-line handwriting recognition is the process of finding letters and words are present in digital image of handwritten text. Several methods of recognition of English, Latin, Arabic, Chinese scripts are excellently reviewed in ([1], [2], [3], [4]). Research in HCR is popular for various practical applications such as reading aid for the blind, bank cheques, automatic pin code reading for sorting of postal mail.

Although many pieces of work have been done on the recognition of printed characters of Indian languages, but only a few attempts have been made towards the recognition of handwritten characters. Most of the research was focused on recognition of off-line handwritten characters for Devanagari and Bangla scripts. It is observed from the literature survey that there is a lot of demand on Indian scripts character recognition and an excellent review has been done on the OCR for Indian languages [5]. A Detailed Study and Analysis of OCR Research on South Indian Scripts is presented in [6].

Rajput and Mali [7] have proposed an efficient method for recognition of isolated Devanagari handwritten numerals based on Fourier descriptors. In [8] zone centroid is computed and the image is further divided in to n equal zones. Average distance from the zone centroid to the each pixel present in the zone is computed. This procedure is repeated for all the zones present in the numeral image. Finally n such features are extracted for classification and recognition. F-ratio Based Weighted Feature Extraction for Similar Shape Character Recognition for different scripts like Arabic/Persian, Devnagari English, Bangla , Oriya, Tamil, Kannada, Telugu etc is presented in [9]. An efficient and novel method for recognition of machine printed and handwritten Kannada numerals using Crack codes and Fourier Descriptors is reported in [10].

Selection of a feature extraction method is an important factor in achieving high recognition performance. A survey of feature extraction methods for character recognition is reported in [11]. Literature survey reveals that the automatic recognition of handwritten numeral has been the subject of intensive research during the last few decades. Numeral identification is very vital in applications such as interpretation of identity numbers, Vehicle registration numbers, Pin Codes, etc. In Indian context, it is evident that still handwritten numeral recognition research is a fascinating area of research to design a robust optical character recognition (OCR), in particular for handwritten Kannada numeral recognition.

Kannada along with other Indian language scripts shares a large number of structural features. Kannada has 49 basic characters, which are classified into three categories: swaras (vowels), vyanjans (consonants) and yogavaahas (part vowel, part consonants). The scripts also include 10 different Kannada numerals of the decimal number system. The rest of the paper is organized as follows: the description of the proposed method is given in section II, experimental results are discussed in section III, Comparative study and conclusion in sections IV and V respectively.

## II. Proposed Methodology

### A. Data collection

To the best of our knowledge standard dataset for handwritten and printed Kannada numerals is not available till today. Therefore, dataset of totally unconstrained handwritten Kannada numerals 0 to 9 is created by collecting the handwritten documents from writers belonging to different professions. The skew in the documents has not been considered. A sample image of scanned document is shown in Fig 1. The individual numerals were extracted manually from the scanned documents and labelled. The labelled numerals are then pre-processed.

Kannada lacks a standard test bed of character images for OCR performance evaluation. To make handwritten Kannada characters font and size independent is a challenge in the research because generalization will be more difficult. The challenging part of Kannada handwritten character recognition is the distinction between the similar shaped components. A very small variation between two characters or numerals leads to recognition complexity and degree of recognition accuracy. The style of writing characters is highly different and they come in various sizes and shapes. Same numeral may take different shapes and conversely two or more different numerals of a script may take similar shape. Some examples of the similar shaped numerals are as shown in figure 2

Fig .1 A sample sheet of Kannada Handwritten numerals 0 to 9

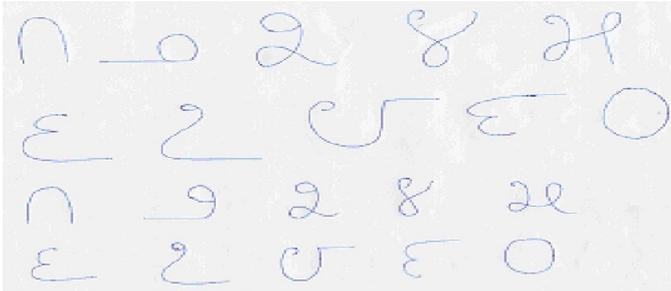

| Numeral 0 and 1 | | |
|---|---|---|
| Numeral 6 and 9 | | |
| Numeral 3 and 7 | | |

Figure 2: Examples of some similar shaped numerals

### B. Feature Extraction

**Method-I**

Histograms of direction chain code of the contour points of the numerals are used as feature for recognition [12]. In this paper dimensional feature extraction technique is used for high speed recognition and is described as below:

**Algorithm**

Begin
Input: a set of pre-processed sample images
Output: a database i.e., a feature matrix of numerals
Method:
Step1: Binary image is Normalized and resized to 30*30 pixels.
Step 2: For an object point in the image, consider a 3*3 window surrounded to the object point. If any one of the four neighbouring points is a background point then this object point (P) is considered as contour point. (As shown in fig 3)
Step 3: The bounding box of an input numeral is divided into 10*10 blocks.
Step 4: In each of these blocks the direction chain code for each contour point is noted and the frequency of the direction codes is computed.
Step 5: Here Chain code of 8 directions is used. Illustration of the same is shown in fig 4 .the chain code used for the different directions is shown in the Table I. Fig 5 shows an example of 8 directional chain code for the numeral '0'.
Step 6: In each block an array of eight integer values representing the frequencies is computed and those frequency values are used as features. Thus for each numeral having 10*10 blocks, (10*10*8=800) 800 features are obtained.
End

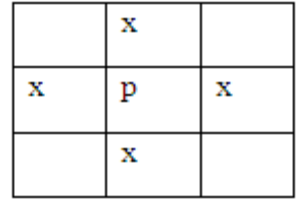

Fig 3: Selecting contour point

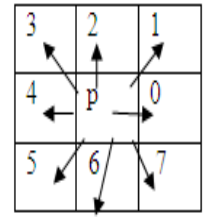

Fig 4: Direction codes for the contour point P.

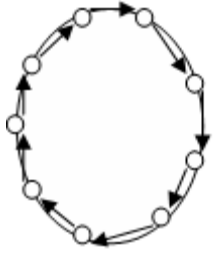

Fig 5:8-directional chain code for number 0: 076553221

TABLE I: CHAIN CODE USED FOR THE DIFFERENT DIRECTIONS

| Directions | Angle in degrees |
|---|---|
| 0 | 0 |
| 1 | 45 |
| 2 | 90 |
| 3 | 135 |
| 4 | 180 |
| 5 | -135 |
| 6 | -90 |
| 7 | -45 |

**Method-II**

In the literature it is observed that most of the character recognition methods focuses on extracting either statistical features such as zoning, moments etc. or the structural features based on the geometry of the character. In this paper we have proposed a method, which attempts to combine both the statistical and structural features. Here we divide the entire image into 9 equal sized zones as indicated in the figure 6

In a binary image, whenever a pixel value changes from 0 to 1 or 1 to 0 it indicates the information about the edge. This information is very significant as it denotes the geometry of the character and helps in identifying the character [13]. In order to capture this information, we have used Run Length Count (RLC) technique. In the proposed method, for every zone, we find the Run Length count in horizontal and vertical direction. A total of 18 features will be extracted for each characters and this will serve as feature vector. The above method is summarized in algorithm

Algorithm
Begin
Input: a set of pre-processed sample images
Output: feature vector for the numerals
Method:
Step1: Binaries the image using a threshold value. The threshold value for an image is fixed using the Otsu's method.
Step 2: The image is resized to 72 * 72 pixels
Step 3: Divide the image into 9 blocks as shown in the figure 7
Step 4: For each block the horizontal and vertical run length is found. Fig 8 illustrates the procedure to find the horizontal run length for a block. Similar approach can be adapted to find the vertical run length. Horizontal run length is 10 in this case
Step 5: A feature vector of length 18 i.e. one horizontal and one vertical run length for each block is obtained for each of the numerals.

End

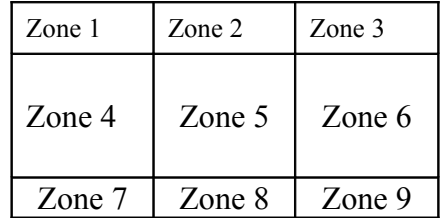

Figure 6: Image Zoning

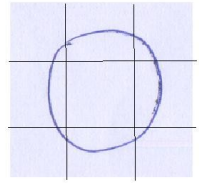

Figure 7: Image Zoning of numeral 0

| 0 | 1 | 1 | 1 | 0 | 0 |
|---|---|---|---|---|---|
| 0 | 0 | 0 | 1 | 1 | 0 |
| 1 | 1 | 1 | 0 | 0 | 0 |
| 0 | 1 | 1 | 0 | 0 | 0 |
| 1 | 1 | 1 | 0 | 0 | 0 |
| 1 | 1 | 0 | 0 | 0 | 0 |
| 0 | 0 | 0 | 0 | 1 | 0 |

Figure 8: Illustration of horizontal run length count

C. *Classifier Fusion*

Over the last decade, ensemble technique is widely used in many different applications. There are different types of ensemble methods. One such type is classification fusion method. In this method, many classifiers are trained on a same feature space. The results of these classifiers are combined to obtain a more accurate classification [14]. In this paper, we have used KNN classifier and the linear classifier. The features obtained from run length coding and from directional chain code are applied to KNN and linear classifiers separately. The prediction of these classifiers is merged using majority-voting technique to correctly classify the sample.

The K-Nearest Neighbor Classifier is an efficient technique to use when the classification problem has pattern classes that display a reasonably limited degree of variability. It considers each input pattern given to it and classifies it to a certain class by calculating the distance between the input pattern and the training patterns. It takes into account only k nearest prototypes to the input pattern during classification. The decision is generally based on the majority of class values of the k nearest neighbors. In the k-Nearest neighbor classification, we compute the distance between features of the test sample and the feature of every training sample.

Linear classifier is a stastical classifier, which makes a classification decision based on the value of the linear combination of the features. A linear classifier is often used in situations where the speed of classification is an issue, since it is often the fastest classifier [12]. Linear classifiers often work very well when the number of dimensions in feature vector is large. It can be represented as shown in equation 1

$$y = f(\vec{w} \cdot \vec{x}) = f\left(\sum_j w_j x_j\right), \quad (1)$$

Where wj is weight vector, learned from a set of labeled training samples.
Xj is the feature vector of testing sample.
f is a simple function that maps the value to the respective classes based on a certain threshold.

The classifier fusion method is summarised in the following algorithm
Algorithm
Begin
Input: a set of training samples and test sample
Output: the class label for which the test sample belongs
Method:

Step1: Extract the directional chain code and run length code features for the training sample and test sample using previously discussed Method-I and Method-II respectively.

Step 2: Apply the features obtained for the training samples to train the KNN and Linear classifiers

Step 3: Apply the features obtained from the test sample to each of the classifier. Let the prediction of the classifiers be p1,p2,p3 and p4

Step 4: Predict the class of the test sample as
    Class = Majority of { p1, p2, p3, p4 }
End
Experimental results
For the experimentation, we have considered a database with nearly 600 samples. Experimentation was conducted using directional chain code features and run length count. For classification, we have used linear classifier and K-nearest neighbor classifier. 80% of the data is used for the training of the classifiers. From the experimentation, we noted that the overall numeral recognition accuracy of the directional features is better than the run length count features. However the run length count feature uses only 18 features as compared with the 72 features of the directional features. For the directional features both the classifiers yielded same average accuracy. The accuracy of the KNN and Linear classifier varied for the run length count feature. We computed accuracy of the individual numerals and observed that for the directional chain code feature, the lowest accuracy was obtained for the numeral 3 in the directional feature method. The common misclassification of numeral 3 was with numeral 7, which is very similar in shape. However the method was able to achieve good recognition rate for the other similar shaped numerals like 0 and 1, 6 and 9. Similarly for the RLC features, the lowest accuracy was obtained for the numeral 9.The common misclassification of numeral 9 was with numeral 2 and 7 even though they are not of similar shape. This may be due to the horizontal strokes present in those numerals. Finally, we applied classifier fusion method for the same dataset and it was found that the overall recognition accuracy was improved when compared to any single classifier, which we had used. The classifier fusion method was able to take the advantage of both the methods. This is reflected in the results. The performance of each classifier with the classifier fusion method is tabulated in Table-II. The comparison of different classifiers with the classifier fusion method is shown in fig.9

TABLE II: RECOGNITION ACCURACY OF THE NUMERALS

| Class of the Numeral | RLC & KNN | RLC & LC | DCC & KNN | DCC & LC | CF |
|---|---|---|---|---|---|
| 0 | 90 | 90 | 100 | 100 | 100 |
| 1 | 90 | 90 | 100 | 100 | 100 |
| 2 | 100 | 100 | 100 | 100 | 100 |
| 3 | 90 | 90 | 70 | 70 | 100 |
| 4 | 100 | 100 | 90 | 100 | 100 |
| 5 | 70 | 90 | 100 | 100 | 90 |
| 6 | 70 | 80 | 100 | 100 | 100 |
| 7 | 100 | 70 | 100 | 90 | 100 |
| 8 | 90 | 90 | 100 | 80 | 90 |
| Avg | 87 | 86 | 94 | 94 | 96 |

III. COMPARATIVE STUDY

The Table III shows the comparison of existing methods with proposed method. Here an attempt is made to achieve good accuracy using less number of features and thereby improving the time and space complexity.

IV. CONCLUSIONS

In this paper, we have implemented two feature extraction method like run length count and directional chain code for the recognition of handwritten Kannada numerals. Initially, K-Nearest Neighbor and Linear classifiers are used for the classification. Further, the recognition accuracy was improved using classifier fusion method. With this approach, we were able to achieve an average recognition rate of 96%. Our major contribution is in the form of reducing the features and testing the effectiveness of classifier fusion method. This method can also be tried for other regional languages. Our method has been compared with several other methods and the accuracies claimed by the various authors have been compared with the accuracy obtained by our methodology. Even though certain authors have claimed that the accuracy obtained is 97%, we have achieved 96% accuracy considering less number of features. Since, our aim is to extract the features, which are dominating for the recognition of numerals, a good amount of accuracy has been achieved with few features. This also improves the complexity of the algorithm

TABLE III: COMPARISION OF PROPOSED METHOD WITH THE EXISTING METHODS

| Authors | No. of samples in data set | Feature extraction method | Classifier | Accuracy (%) |
|---|---|---|---|---|
| B V Dhandra et al[15] | 1512 | Structural features | KNN classifer with $2^{nd}$ order weighted Minkowski measure | 96.12 |
| Rajashekhara Aaradhya S V et.al[16] | 1000 | Vertical projection distance with zoning | NN classifier | 93 |
| R Sanjeev Kunte et al[17] | 2500 | Wavelet | Nueral classifier | 92.3 |
| G G Rajput et al[18] | 1000 | Image fusion | NN classifier | 91.2 |
| V N Manjunath Aaradhya et al[19] | 2000 | Radon features | NN classifier | 91.2 |
| Dinesh Aacharya U et. al[20] | 500 | Structural features | K-means | 90.5 |
| Rajashekhara Aaradhya S V et.al[21] | 4000 | Zoning with average angle from image centroid to each pixel in the zone | SVM | 97.85 |
| G G Rajput et al[22] | 2500 | Chain code an fourier descripter | SVM | 97.34 |
| Proposed Method | 600 | Directional Features & RLC | KNN and Linear classifier fusion | 96 |

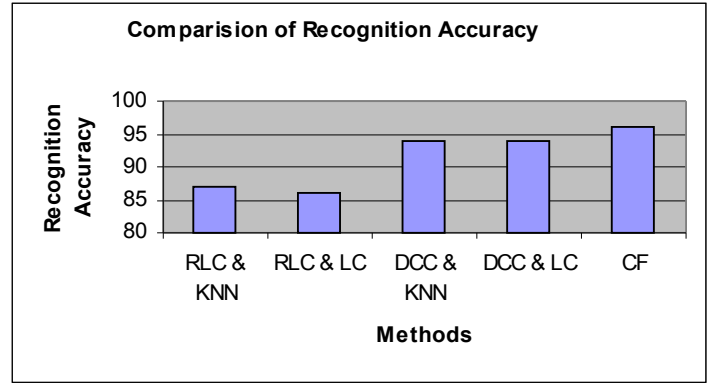

Figure 9: Comparison of various methods